\documentclass[runningheads]{llncs}

\usepackage[dvips]{graphicx}

\usepackage{amsmath,amssymb} 

\usepackage{color}

\usepackage{footnote}

\definecolor{mygray}{gray}{0.4}

\usepackage[width=122mm,left=12mm,paperwidth=146mm,height=193mm,top=12mm,paperheight=217mm]{geometry}

\usepackage{array}

\usepackage{wrapfig}

\usepackage[numbers,sort]{natbib}

\newcommand{\eat}[1]{}

\begin{document}

\pagestyle{headings}

\mainmatter

\def\ECCV16SubNumber{1641}  

\newcommand{\name}{stochastic depth}

\title{Deep Networks with Stochastic Depth}

\titlerunning{Deep Networks with Stochastic Depth}

\authorrunning{Gao Huang*, Yu Sun*, Zhuang Liu, Daniel Sedra, Kilian Q. Weinberger}

\author{{Gao Huang*, Yu Sun*, Zhuang Liu$^\dagger$, Daniel Sedra, Kilian Q. Weinberger}}
\institute{
* Authors contribute equally\\
\url{\texttt{\{gh349, ys646, dms422, kqw4\}@cornell.edu}}, Cornell University\\
$^\dagger$ \url{\texttt{liuzhuang13@mails.tsinghua.edu.cn}}, Tsinghua University
}

\maketitle

\begin{abstract}

Very deep convolutional networks with hundreds of layers have led to significant reductions in error on competitive benchmarks. Although the unmatched expressiveness of the many layers can be highly desirable at test time, training very deep networks comes with its own set of challenges. The gradients can vanish, the forward flow often diminishes, and the training time can be painfully slow. To address these problems, we propose \emph{\name{}}, a training procedure that enables the seemingly contradictory setup to \emph{train short} networks and \emph{use deep} networks at test time. We start with very deep networks but during training, for each mini-batch, randomly drop a subset of layers and bypass them with the identity function. This simple approach complements the recent success of residual networks. It reduces training time substantially and improves the test error significantly on almost all data sets that we used for evaluation. With \name{} we can increase the depth of residual networks even beyond 1200 layers and still yield meaningful improvements in test error (4.91\% on CIFAR-10).


\end{abstract}

\section{Introduction}

\label{sec_intro}


Convolutional Neural Networks (CNNs) were arguably popularized within the vision community in 2009 through AlexNet~\cite{krizhevsky2009learning} and its celebrated victory at the ImageNet competition~\cite{deng2009imagenet}.
Since then there has been a notable shift towards CNNs in many areas of computer vision~\cite{krizhevsky2012imagenet,sermanet2013overfeat,simonyan2014very,springenberg2014striving,szegedy2015going,he2015deep}. As this shift unfolds, a second trend emerges; deeper and deeper CNN architectures are being developed and trained. Whereas AlexNet had 5 convolutional layers~\cite{krizhevsky2009learning}, the VGG network and GoogLeNet in 2014 had 19 and 22 layers respectively~\cite{simonyan2014very,szegedy2015going}, and most recently the ResNet architecture featured 152 layers~\cite{he2015deep}.

Network depth is a major determinant of model expressiveness, both in theory \cite{haastad1991power,haastad1987computational} and in practice \cite{he2015deep,simonyan2014very,szegedy2015going}. However, very deep models also introduce new challenges: vanishing gradients in backward propagation, diminishing feature reuse in forward propagation, and long training time.

\emph{Vanishing Gradients} is a well known nuisance in neural networks with many layers~\cite{bengio1994learning}. As the gradient information is back-propagated, repeated multiplication or convolution with small weights renders the gradient information ineffectively small in earlier layers. Several approaches exist to reduce this effect in practice, for example through careful initialization \cite{glorot2010understanding}, hidden layer supervision \cite{lee2014deeply}, or, recently, Batch Normalization~\cite{ioffe2015batch}.

\emph{Diminishing feature reuse} during forward propagation (also known as loss in information flow \cite{srivastava2015highway}) refers to the analogous problem to vanishing gradients in the forward direction. The features of the input instance, or those computed by earlier layers, are ``washed out'' through repeated multiplication or convolution with (randomly initialized) weight matrices, making it hard for later layers to identify and learn ``meaningful'' gradient directions. Recently, several new architectures attempt to circumvent this problem through direct identity mappings between layers, which allow the network to pass on features unimpededly from earlier layers to later layers~\cite{he2015deep,srivastava2015highway}.

\emph{Long training time} is a serious concern as networks become very deep. The forward and backward passes scale linearly with the depth of the network. Even on modern computers with multiple state-of-the-art GPUs, architectures like the 152-layer ResNet require several weeks to converge on the ImageNet dataset~\cite{he2015deep}.

The researcher is faced with an inherent dilemma:
shorter networks have the advantage that information flows efficiently forward and backward, and can therefore be trained effectively and within a reasonable amount of time. However, they are not expressive enough to represent the complex concepts that are commonplace in computer vision applications. Very deep networks have much greather model complexity, but are very difficult to train in practice and require a lot of time and patience.

In this paper, we propose \emph{deep networks with \name{}}, a novel training algorithm that is based on the seemingly contradictory insight that ideally we would like to have a \emph{deep} network during \emph{testing} but a \emph{short} network during \emph{training}. We resolve this conflict by creating deep Residual Network~\cite{he2015deep} architectures (with hundreds or even thousands of layers) with sufficient modeling capacity; however, during training  we shorten the network significantly by randomly removing a substantial fraction of layers independently for each sample or mini-batch.
The effect is a network with a small \emph{expected} depth during training, but a large depth during testing. Although seemingly simple, this approach is surprisingly effective in practice.

In extensive experiments we observe that training with \name{} substantially reduces training time and test error (resulting in multiple new records to the best of our knowledge at the time of initial submission to ECCV).
The reduction in training time can be attributed to the shorter forward and backward propagation, so the training time no longer scales with the full depth, but the shorter \emph{expected depth} of the network. We attribute the reduction in test error to two factors: 1) shortening the (expected) depth during training reduces the chain of forward propagation steps and gradient computations, which strengthens the gradients especially in earlier layers during backward propagation; 2) networks trained with \name{} can be interpreted as an implicit \emph{ensemble} of networks of different depths, mimicking the record breaking ensemble of depth varying ResNets trained by He et al.~\cite{he2015deep}.

We also observe that similar to Dropout \cite{srivastava2014dropout}, training with \name{} acts as a regularizer, even in the presence of Batch Normalization \cite{ioffe2015batch}. On experiments with CIFAR-10, we increase the depth of a ResNet beyond 1000 layers and still obtain significant improvements in test error.

\section{Background}

\label{sec_related}


Many attempts have been made to improve the training of very deep networks. Earlier works adopted greedy layer-wise training or better initialization schemes
to alleviate the vanishing gradients and diminishing feature reuse problems \cite{fahlman1989cascade,erhan2010does,glorot2010understanding}.
A notable recent contribution towards training of very deep networks is Batch Normalization~\cite{ioffe2015batch}, which standardizes the mean and variance of hidden layers with respect to each mini-batch. This approach reduces the vanishing gradients problem and yields a strong regularizing effect.

Recently, several authors introduced extra skip connections to improve the information flow during forward and backward propagation. Highway Networks \cite{srivastava2015highway} allow earlier representations to flow unimpededly to later layers through parameterized skip connections known as ``information highways'', which can cross several layers at once. The skip connection parameters, learned during training, control the amount of information allowed on these ``highways''.

\paragraph{\textbf{Residual networks (ResNets)}\cite{he2015deep}} simplify Highway Networks by shortcutting (mostly) with identity functions. This simplification greatly improves training efficiency, and enables more direct feature reuse. ResNets are motivated by the observation that neural networks tend to obtain \emph{higher training error} as the depth increases to very large values. This is counterintuitive, as the network gains more parameters and therefore better function approximation capabilities. The authors conjecture that the networks become \emph{worse} at function approximation because the gradients and training signals vanish when they are propagated through many layers. As a fix, they propose to add \emph{skip connections} to the network. Formally, if $H_\ell$ denotes the output of the $\ell^{th}$ layer (or sequence of layers) and $f_\ell(\cdot)$ represents a\ typical convolutional transformation from layer $\ell\!-\!1$ to $\ell$, we obtain
\begin{equation}
\label{eq:res_block}
 H_{\ell} = {\tt ReLU} (f_\ell( H_{\ell-1}) + \text{id} (H_{\ell-1})),
\end{equation}
where $\text{id}(\cdot)$ denotes the identity transformation and we assume a ${\tt ReLU}$ transition function~\cite{nair2010rectified}. Fig.~\ref{fig:resblock} illustrates an example of a function $f_\ell$, which consists of multiple convolutional and Batch Normalization layers.
When the output dimensions of $f_\ell$ do not match those of $H_{\ell-1}$, the authors redefine $\text{id}(\cdot)$ as a linear projection to reduce the dimensions of $\text{id}(H_{\ell-1})$ to match those of $f_\ell(H_{\ell-1})$.
The propagation rule in \eqref{eq:res_block} allows the network to pass gradients and features (from the input or those learned in earlier layers) back and forth between the layers via the identity transformation $\text{id}(\cdot)$.

\paragraph{\textbf{Dropout.}} Stochastically dropping hidden nodes or connections has been a popular regularization method for neural networks. The most notable example is Dropout \cite{srivastava2014dropout}, which multiplies each hidden activation by an independent Bernoulli random variable. Intuitively, Dropout reduces the effect known as ``co-adaptation'' of hidden nodes collaborating in groups instead of independently producing useful features; it also makes an analogy with training an ensemble of exponentially many small networks. Many follow up works have been empirically successful, such as DropConnect \cite{icml2013_wan13}, Maxout \cite{goodfellow2013maxout} and DropIn \cite{smith2015gradual}.

Similar to Dropout, \name{} can be interpreted as training an ensemble of networks, but with different depths, possibly achieving higher diversity among ensemble members than ensembling those with the same depth. Different from Dropout, we make the network shorter instead of thinner, and are motivated by a different problem. Anecdotally, Dropout loses effectiveness when used in combination with Batch Normalization~\cite{ioffe2015batch,blogpostcifar10}. Our own experiments with various Dropout rates (on CIFAR-10) show that Dropout gives practically no improvement when used on 110-layer ResNets with Batch Normalization.

We view all of these previous approaches to be extremely valuable and consider our proposed training with \name{} complimentary to these efforts. In fact, in our experiments we show that training with \name{} is indeed very effective on ResNets with Batch Normalization.

\section{Deep Networks with Stochastic Depth}

\label{sec_method}


\begin{figure*}[t]
	\begin{center}
		\vspace{-2ex}
		\centerline{\includegraphics[width=1\columnwidth]{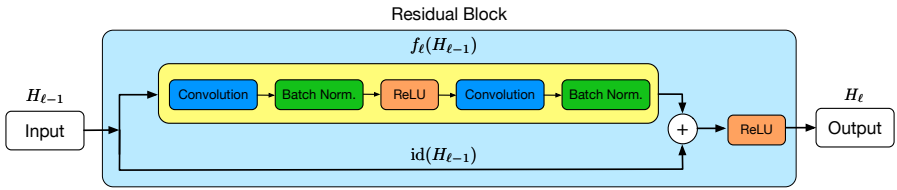}}
		\vspace{-2ex}
		\caption{A close look at the $\ell^\text{th}$ ResBlock in a ResNet.\label{fig:resblock}}
		\vspace{-6ex}
	\end{center}
\end{figure*}

Learning with \name{} is based on a simple intuition. To reduce the \emph{effective} length of a neural network during training, we randomly skip layers entirely. We achieve this by introducing skip connections in the same fashion as ResNets, however the connection pattern is randomly altered for each mini-batch. For each mini-batch we randomly select sets of layers and remove their corresponding transformation functions, only keeping the identity skip connection.
Throughout, we use the architecture described by He et al.~\cite{he2015deep}. Because the architecture already contains skip connections, it is straightforward to modify, and isolates the benefits of \name{} from that of the ResNet identity connections. Next we describe this network architecture and then explain the \name{} training procedure in detail.

\paragraph{\textbf{ResNet architecture.}}
Following He et al.~\cite{he2015deep}, we construct our network as the functional composition of $L$ \emph{residual blocks} (ResBlocks), each encoding the update rule~\eqref{eq:res_block}.  Fig.~\ref{fig:resblock} shows a schematic illustration of the $\ell^{th}$ ResBlock. In this example, $f_\ell$ consists of a sequence of layers: {\tt Conv-BN-ReLU-Conv-BN}, where {\tt Conv} and {\tt BN} stand for Convolution and Batch Normalization respectively.
This construction scheme is adopted in all our experiments except ImageNet, for which we use the bottleneck block detailed in He et al. \cite{he2015deep}. Typically, there are 64, 32, or 16 filters in the convolutional layers (see Section~\ref{sec_results} for experimental details).

\begin{figure*}[t]
	\begin{center}
		\vspace{-2ex}
		\centerline{\includegraphics[width=1\columnwidth]{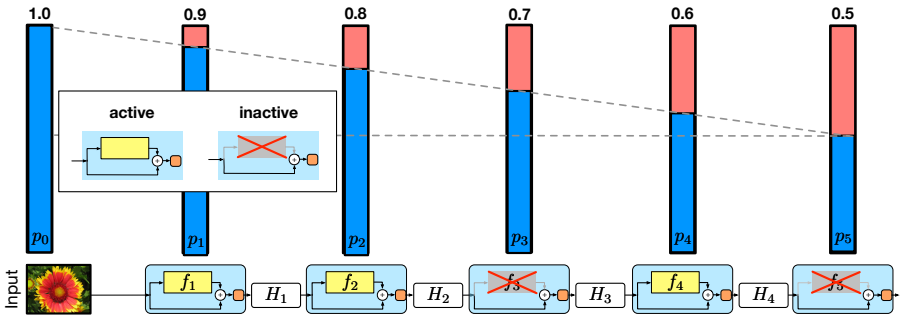}}
		\caption{The linear decay of $p_\ell$ illustrated on a ResNet with \name{} for $p_0\!=\!1$ and $p_L\!=\! 0.5$. Conceptually, we treat the input to the first ResBlock as $H_0$, which is always active.}
    \label{figure.lin_decay}
		\vspace{-6ex}
	\end{center}
\end{figure*}

\paragraph{\textbf{Stochastic depth}} aims to shrink the depth of a network during training, while keeping it unchanged during testing. We can achieve this goal by randomly dropping entire ResBlocks during training and bypassing their transformations through skip connections.
Let $b_\ell\in\{0,1\}$ denote a Bernoulli random variable, which indicates whether the $\ell^{th}$ ResBlock is active ($b_\ell=1$) or inactive ($b_\ell=0$). Further, let us denote the ``survival'' probability of ResBlock $\ell$ as $p_\ell=\Pr(b_\ell=1)$.

With this definition we can bypass the $\ell^{th}$ ResBlock by multiplying its function $f_\ell$ with $b_\ell$ and we extend the update rule from~\eqref{eq:res_block} to
\begin{equation}
\label{res_block_drop}
 H_\ell = {\tt ReLU} (b_\ell f_\ell(H_{\ell-1}) +  \text{id}(H_{\ell-1})).
\end{equation}
If $b_\ell\!=\!1$, eq.~\eqref{res_block_drop} reduces to the original ResNet update \eqref{eq:res_block} and this ResBlock remains unchanged.
If $b_\ell\!=\!0$, the ResBlock reduces to the identity function,
\begin{equation}
 H_\ell = \text{id}(H_{\ell-1}).
\end{equation}
This reduction follows from the fact that the input $H_{\ell-1}$ is always non-negative, at least for the architectures we use. For $\ell\geq 2$, it is the output of the previous ResBlock, which is non-negative because of the final ${\tt ReLU}$ transition function (see Fig.~\ref{fig:resblock}). For $\ell\!=\!1$, its input is the output of a {\tt Conv-BN-ReLU} sequence that begins the architecture before the first ResBlock. For non-negative inputs the ${\tt ReLU}$ transition function acts as an identity.

\paragraph{\textbf{The survival probabilities}} $p_\ell$ are new hyper-parameters of our training procedure. Intuitively, they should take on similar values for neighboring ResBlocks. One option is to set $p_\ell=p_L$ uniformly for all $\ell$ to obtain a single hyper-parameter $p_L$. Another possibility is to set them according to a smooth function of $\ell$. We propose a simple linear decay rule from $p_0=1$ for the input, to $p_L$ for the last ResBlock:
\begin{equation}
\label{eq:surv_rate}
 p_\ell = 1-\frac{\ell}{L}(1-p_L).
\end{equation}
See Fig.~\ref{figure.lin_decay} for a schematic illustration.
The linearly decaying survival probability originates from our intuition that the earlier layers extract low-level features that will be used by later layers and should therefore be more reliably present.
In Section~\ref{sec_results} we perform a more detailed empirical comparison between the uniform and decaying assignments for $p_\ell$.  We conclude that the linear decay rule \eqref{eq:surv_rate} is preferred and, as training with \name{} is surprisingly stable with respect to $p_L$, we set $p_L=0.5$ throughout (see Fig.~\ref{figure.deathRate}).

\paragraph{\textbf{Expected network depth}.}
During the forward-backward pass the transformation $f_\ell$ is bypassed with probability $(\!1-p_\ell\!)$, leading to a network with reduced depth.
With \name{}, the number of effective ResBlocks during training, denoted as $\tilde{L}$, becomes a random variable. Its expectation is given by:
$
E(\tilde{L}) =  \sum\nolimits_{\ell=1}^L p_\ell.
$

Under the linear decay rule with $p_L=0.5$, the expected number of ResBlocks during training reduces to $E(\tilde{L})=(3L-1)/4$, or $E(\tilde{L})\approx 3L/4$ when $L$ is large. For the 110-layer network with $L=54$ commonly used in our experiments, we have $E(\tilde{L})\approx 40$. In other words, with \name{}, we train ResNets with an average number of 40 ResBlocks, but recover a ResNet with 54 blocks at test time.
This reduction in depth significantly alleviates the vanishing gradients and the information loss problem in deep ResNets. Note that because the connectivity is random, there will be updates with significantly shorter networks and more direct paths to individual layers. We provide an empirical demonstration of this effect in Section \ref{sec_analysis}.

\paragraph{\textbf{Training time savings}.}
When a ResBlock is bypassed for a specific iteration, there is no need to perform forward-backward computation or gradient updates. As the forward-backward computation dominates the training time, \name{} significantly speeds up the training process. Following the calculations above, approximately $25\%$ of training time could be saved under the linear decay rule with $p_L=0.5$. The timings in practice using our implementation are consistent with this analysis (see the last paragraph of Section~\ref{sec_results}). More computational savings can be obtained by switching to a uniform probability for $p_\ell$ or lowering $p_L$ accordingly. In fact, Fig.~\ref{figure.deathRate} shows that with $p_L=0.2$, the ResNet with \name{} obtains the same test error as its constant depth counterpart on CIFAR-10 but gives a 40\% speedup.

\paragraph{\textbf{Implicit model ensemble}.}
In addition to the predicted speedups, we also observe significantly lower testing errors in our experiments, in comparison with ResNets of constant depth.
One explanation for our performance improvements is that training with \name{} can be viewed as training an ensemble of ResNets \emph{implicitly}.
Each of the $L$ layers is either active or inactive, resulting in $2^L$ possible network combinations. For each training mini-batch one of the $2^L$ networks (with shared weights) is sampled and updated. During testing all networks are averaged using the approach in the next paragraph.

\paragraph{\textbf{Stochastic depth during testing}} requires small modifications to the network. We keep all functions $f_\ell$ active throughout testing in order to utilize the full-length network with all its model capacity. However, during training, functions $f_\ell$ are only active for a fraction $p_\ell$ of all updates, and the corresponding weights of the next layer are calibrated for this survival probability. We therefore need to re-calibrate the outputs of any given function $f_\ell$ by the expected number of times it participates in training, $p_\ell$. The forward propagation update rule becomes:
\begin{equation}
\label{res_block_drop_test}
 H_\ell^\text{Test} = {\tt ReLU} (p_\ell f_\ell(H_{\ell-1}^\text{Test}; W_\ell) +  H_{\ell-1}^\text{Test}).
\end{equation}
From the model ensemble perspective, the update rule~\eqref{res_block_drop_test} can be interpreted as combining all possible networks into a single test architecture, in which each layer is weighted by its survival probability.

\section{Results}

\label{sec_results}

\begin{table}[t!]
\vskip -0.05in
\caption{Test error (\%) of ResNets trained with \name{} compared to other most competitive methods previously published (whenever available). A "+" in the name denotes standard data augmentation. ResNet with constant depth refers to our reproduction of the experiments by He et al.}
\label{table.error}
\vspace{-2ex}
\begin{center}
\resizebox{0.9\textwidth}{!}
{
\begin{small}
\begin{tabular}{ccccc}
	
\hline
		  	& CIFAR10+   & CIFAR100+   & SVHN & ImageNet\\
\hline
Maxout \cite{goodfellow2013maxout} 		
			& 9.38 	   & -		   & 2.47 	&	-	\\
DropConnect \cite{icml2013_wan13}
	   		& 9.32 	   & -		   & 1.94 	&	-	\\
Net in Net \cite{lin2013network}
			& 8.81 	   & -		   & 2.35 	&	-	\\
Deeply Supervised \cite{lee2014deeply}
			& 7.97 	   & -		   & 1.92 	& 33.70	\\
Frac. Pool \cite{graham2014fractional}
			& - 	   & 27.62	   & -	   	&	-	\\
All-CNN \cite{springenberg2014striving}
			& 7.25 	   & -		   & -	   	& 41.20	\\
Learning Activation \cite{agostinelli2014learning}
			& 7.51 	   & 30.83	   & - 		&   -	\\
R-CNN \cite{liang2015recurrent}
			& 7.09 	   & -		   & 1.77 	&   -	\\
Scalable BO \cite{snoek2015scalable}
			& 6.37 	   & 27.40	   & 1.77 	&	-	\\
Highway Network \cite{srivastava2015training}
			& 7.60	   & 32.24	   & - 		&   -	\\
Gen. Pool \cite{lee2015generalizing}
			& 6.05	   & -		   & 1.69 	& 28.02 \\
\hline
ResNet with constant depth		
		    & 6.41 	   & 27.76	   & 1.80	& 21.78  \\
ResNet with \name
			& 5.25	   & 24.98	   & 1.75   & 21.98  \\
\hline
\end{tabular}
\end{small}
}
\end{center}
\vspace{-4ex}
\end{table} 

We empirically demonstrate the effectiveness of \name{} on a series of benchmark data sets: CIFAR-10, CIFAR-100~\cite{krizhevsky2009learning}, SVHN~\cite{netzer2011reading}, and ImageNet~\cite{deng2009imagenet}.

\paragraph{\textbf{Implementation details.}}
For all data sets we compare the results of ResNets with our proposed \name{} and the original constant depth, and other most competitive benchmarks. We set $p_\ell$ with the linear decay rule of $p_0\!=\!1$ and $p_L\!=\!0.5$ throughout. In all experiments we report the test error from the epoch with the lowest validation error.
For best comparisons we use the same construction scheme (for constant and \name{}) as described by He et al.~\cite{he2015deep}. In the case of CIFAR-100 we use the same 110-layer ResNet used by He et al. \cite{he2015deep} for CIFAR-10, except that the network has a 100-way softmax output. Each model contains three groups of residual blocks that differ in number of filters and feature map size, and each group is a stack of 18 residual blocks. The numbers of filters in the three groups are 16, 32 and 64, respectively. For the transitional residual blocks, i.e. the first residual block in the second and third group, the output dimension is larger than the input dimension. Following He et al. \cite{he2015deep}, we replace the identity connections in these blocks by an average pooling layer followed by zero paddings to match the dimensions.
Our implementations are in Torch 7 \cite{collobert2011torch7}. The code to reproduce the results is publicly available on GitHub at {\tt https://github.com/yueatsprograms/Stochastic\char`_Depth}.

\begin{figure*}[t]
	\begin{center}
		\centerline{
			\includegraphics[width=0.5\columnwidth]{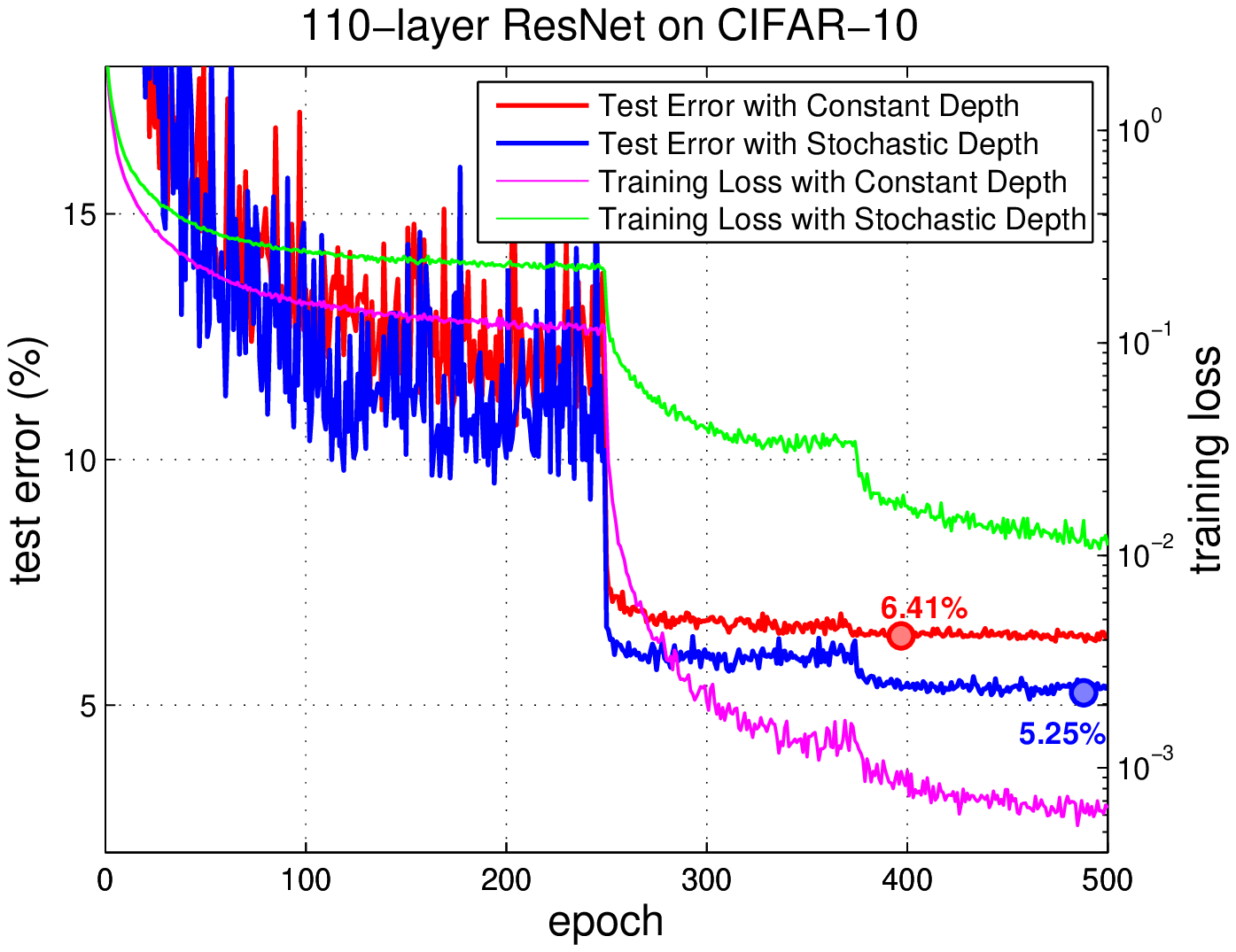}
			\includegraphics[width=0.5\columnwidth]{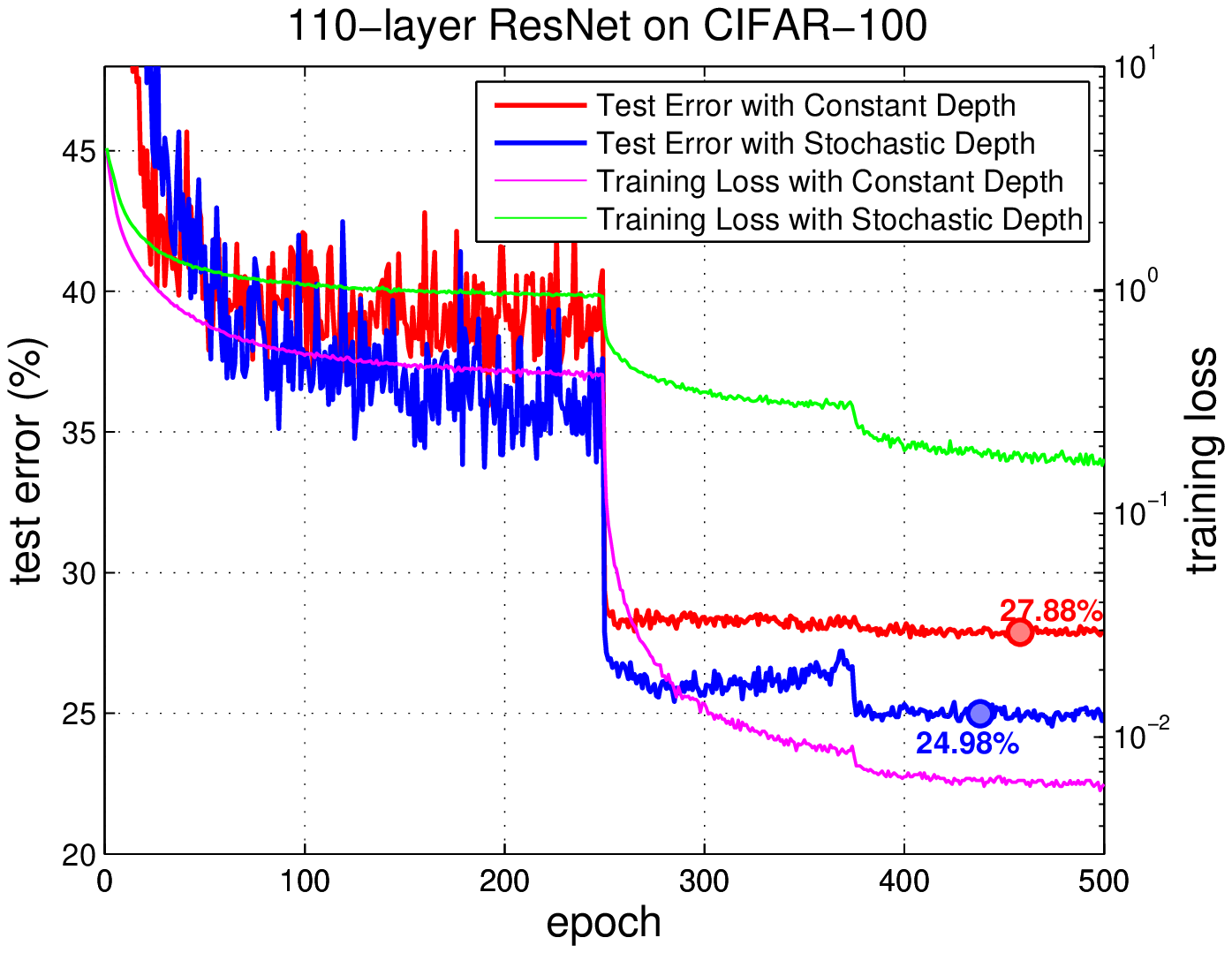}}
		\vspace{-2ex}
		\caption{Test error on CIFAR-10 (\textit{left}) and CIFAR-100 (\textit{right}) during training,  with data augmentation, corresponding to results in the first two columns of Table~\ref{table.error}.}
		\vspace{-8ex}
		\label{figure.cifar}
	\end{center}
\end{figure*}

\paragraph{\textbf{CIFAR-10.}}
CIFAR-10 \cite{krizhevsky2009learning} is a dataset of 32-by-32 color images, representing 10 classes of natural scene objects. The training set and test set contain 50,000 and 10,000 images, respectively. We hold out 5,000 images as validation set, and use the remaining 45,000 as training samples. Horizontal flipping and translation by 4 pixels are the two standard data augmentation techniques adopted in our experiments, following the common practice \cite{goodfellow2013maxout,icml2013_wan13,lin2013network,lee2014deeply,springenberg2014striving,agostinelli2014learning,lee2015generalizing}.

The baseline ResNet is trained with SGD for 500 epochs, with a mini-batch size 128. The initial learning rate is 0.1, and is divided by a factor of 10 after epochs 250 and 375. We use a weight decay of 1e-4, momentum of 0.9, and Nesterov momentum \cite{sutskever2013importance} with 0 dampening, as suggested by \cite{blogpostresnet}. For \name{}, the network structure and all optimization settings are exactly the same as the baseline. All settings were chosen to match the setup of He et al.~\cite{he2015deep}.

The results are shown in Table~\ref{table.error}. ResNets with constant depth result in a competitive 6.41\% error on the test set. ResNets trained with \name{} yield a further relative improvement of $18\%$ and result in 5.25\% test error.
To our knowledge this is significantly lower than the best existing single model performance ($6.05\%$)~\cite{lee2015generalizing} on CIFAR-10 prior to our submission, without resorting to massive data augmentation~\cite{graham2014fractional,springenberg2014striving}.\footnote{The only model that performs even better is the 1202-layer ResNet with \name{}, discussed later in this section. }
Fig.~\ref{figure.cifar} (\textit{left}) shows the test error as a function of epochs. The point selected by the lowest validation error is circled for both approaches. We observe that ResNets with \name{} yield lower test error but also slightly higher fluctuations (presumably due to the random depth during training).

\paragraph{\textbf{CIFAR-100.}}
Similar to CIFAR-10, CIFAR-100~\cite{krizhevsky2009learning} contains 32-by-32 color images with the same train-test split, but from 100 classes. For both the baseline and our method, the experimental settings are exactly the same as those of CIFAR-10. The constant depth ResNet yields a test error of 27.22\%, which is already the state-of-the-art in CIFAR-100 with standard data augmentation. Adding \name{} drastically reduces the error to 24.98\%, and is again the best published single model performance to our knowledge (see Table~\ref{table.error} and Fig.~\ref{figure.cifar} \textit{right}).

We also experiment with CIFAR-10 and CIFAR-100 without data augmentation. ResNets with constant depth obtain 13.63\% and 44.74\% on CIFAR-10 and CIFAR-100 respectively. Adding \name{} yields consistent improvements of about 15\% on both datasets, resulting in test errors of 11.66\%  and 37.8\% respectively.

\begin{figure*}[t]
	\begin{center}
		\centerline{\includegraphics[width=0.5\columnwidth]{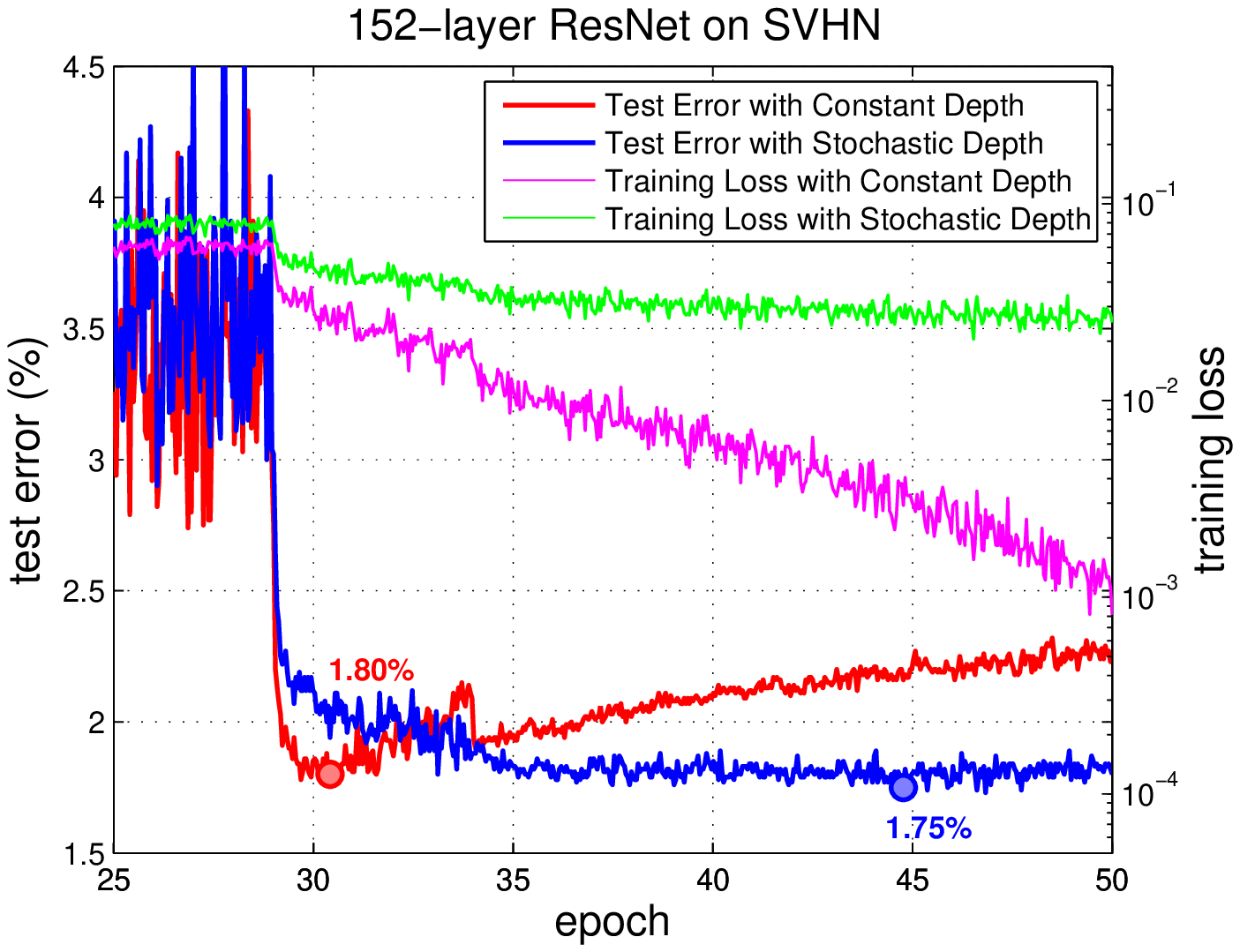} \includegraphics[width=0.5\columnwidth]{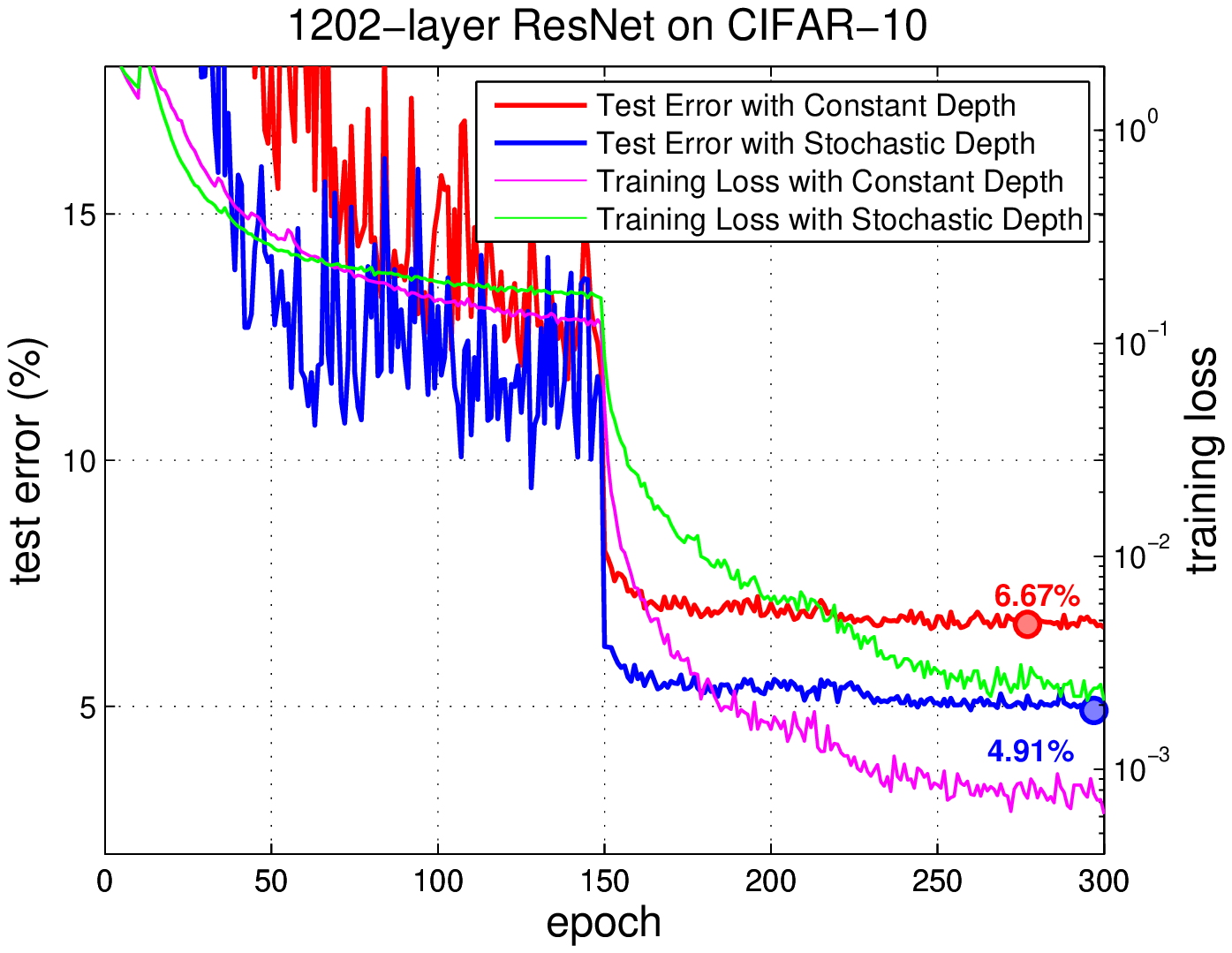}}
		\caption{Left: Test error on SVHN, corresponding to results on column three in Table~\ref{table.error}. \textit{right}: Test error on CIFAR-10 using 1202-layer ResNets. The points of lowest validation errors are highlighted in each case.}
		\vspace{-8ex}
		\label{figure.svhn_cifar1202}
	\end{center}
\end{figure*}

\paragraph{\textbf{SVHN.}}
The format of the Street View House Number (SVHN)~\cite{netzer2011reading} dataset that we use contains 32-by-32 colored images of cropped out house numbers from Google Street View. The task is to classify the digit at the center. There are 73,257 digits in the training set, 26,032 in the test set and 531,131 easier samples for additional training. Following the common practice, we use all the training samples but do not perform data augmentation. For each of the ten classes, we randomly select 400 samples from the training set and 200 from the additional set, forming a validation set with 6,000 samples in total. We preprocess the data by subtracting the mean and dividing the standard deviation. Batch size is set to 128, and validation error is calculated every 200 iterations.

Our baseline network has 152 layers. It is trained for 50 epochs with a beginning learning rate of 0.1, divided by 10 after epochs 30 and 35. The depth and learning rate schedule are selected by optimizing for the validation error of the baseline through many trials. This baseline obtains a competitive result of 1.80\%. However, as seen in Fig.~\ref{figure.svhn_cifar1202}, it starts to overfit at the beginning of the second phase with learning rate 0.01, and continues to overfit until the end of training. With \name{}, the error improves to 1.75\%, the second-best published result on SVHN to our knowledge after~\cite{lee2015generalizing}.

\paragraph{\textbf{Training time comparison.}}

\begin{table}[t!]
\label{table.training_time}
\caption{Training time comparison on benchmark datasets.}
\begin{center}
\resizebox{0.65\textwidth}{!}
{
\begin{small}
\begin{tabular}{lccc}
\hline
		  	& CIFAR10+   & CIFAR100+   & SVHN	\\
\hline
Constant Depth		
		    & 20h 42m    & 20h 51m	   & 33h 43m \\
Stochastic Depth
			& 15h 7m	 & 15h 20m	   & 25h 33m \\
\hline
\end{tabular}
\end{small}
}
\vspace{-5ex}
\end{center}
\end{table} 
We compare the training efficiency of the constant depth and \name{} ResNets used to produce the previous results. Table 2 shows the training (clock) time under both settings with the linear decay rule $p_L=0.5$. Stochastic depth consistently gives a 25\% speedup, which confirms our analysis in Section~\ref{sec_method}. See Fig.~\ref{figure.deathRate} and the corresponding section on hyper-parameter sensitivity for more empirical analysis. 

\begin{wrapfigure}{r}{0.5\textwidth}
	\vspace{-15ex}
	\begin{center}
		\includegraphics[width=0.45\textwidth]{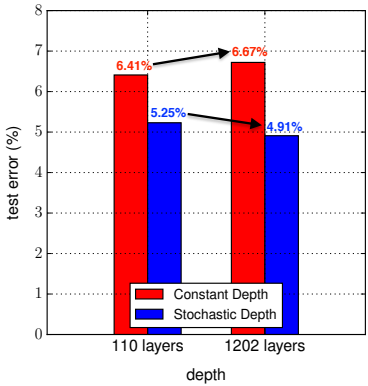}
	\end{center}
	\vspace{-10ex}
	\caption{With \name{}, the 1202-layer ResNet still significantly improves over the 110-layer one.}
	\vspace{-4ex}
	\label{figure.1202compare}
\end{wrapfigure}

\paragraph{\textbf{Training with a 1202-layer ResNet.}}
He et al.~\cite{he2015deep} tried to learn CIFAR-10 using an aggressively deep ResNet with 1202 layers. As expected, this extremely deep network overfitted to the training set: it ended up with a test error of 7.93\%, worse than their 110-layer network. We repeat their experiment on the same 1202-layer network, with constant and stochastic depth. We train for 300 epochs, and set the learning rate to 0.01 for the first 10 epochs to ``warm-up" the network and facilitate initial convergence, then restore it to 0.1, and divide it by 10 at epochs 150 and 225.

The results are summarized in Fig.~\ref{figure.svhn_cifar1202} (\textit{right}) and Fig.~\ref{figure.1202compare}. Similar to He et al.~\cite{he2015deep}, the ResNets with constant depth of 1202 layers yields a test error of 6.67\%, which is worse than the 110-layer constant depth ResNet. In contrast, if trained with \name{}, this extremely deep ResNet performs remarkably well. We want to highlight two trends: 1) Comparing the two 1202-layer nets shows that training with \name{} leads to a 27\% relative improvement; 2) Comparing the two networks trained with \name{}  shows that increasing the architecture from 110 layers to 1202 yields a further improvement on the previous record-low 5.25\%, to a 4.91\% test error without sign of overfitting, as shown in Fig.~\ref{figure.svhn_cifar1202} (\textit{right}) \footnote{We do not include this result in Table \ref{table.error} since this architecture was only trained on one of the datasets.}.

To the best of our knowledge, this is the lowest known test error on CIFAR-10 with moderate image augmentation and the first time that a network with more than 1000 layers has been shown to \emph{further reduce} the test error \footnote{This is, until early March, 2016, when this paper was submitted to ECCV. Many new developments have further decreased the error on CIFAR-10 since then (and some are based on this work).}. We consider these findings highly encouraging and hope that training with \name{} will enable researchers to leverage extremely deep architectures in the future.

\begin{wrapfigure}{t}{0.5\textwidth}
	\begin{center}
		\vspace{-8ex}
		\centerline{\includegraphics[width=0.5\columnwidth]{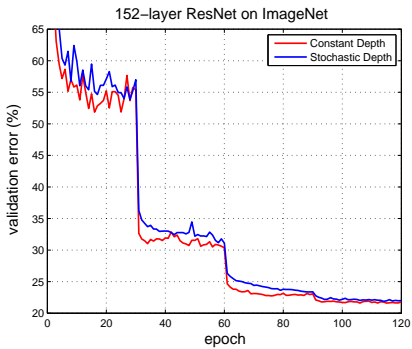}}
		\vspace{-2ex}
		\caption{Validation error on ILSVRC 2012 classification.}
		\vspace{-8ex}
		\label{figure.imagenet}
	\end{center}
\end{wrapfigure}

\paragraph{\textbf{ImageNet.}}
The ILSVRC 2012 classification dataset consists of 1000 classes of images, in total 1.2 million for training, 50,000 for validation, and 100,000 for testing. Following the common practice, we only report the validation errors. We follow He et al.~\cite{he2015deep} to build a 152-layer ResNet with 50 bottleneck residual blocks. When input and output dimensions do not match, the skip connection uses a learned linear projection for the mismatching dimensions, and an identity transformation for the other dimensions. Our implementation is based on the github repository {\tt fb.resnet.torch\footnote{\tt https://github.com/facebook/fb.resnet.torch}}~\cite{blogpostresnet}, and the optimization settings are the same as theirs, except that we use a batch size of 128 instead of 256 because we can only spread a batch among 4 GPUs (instead of 8 as they did).

We train the constant depth baseline for 90 epochs (following He et al. and the default setting in the repository) and obtain a final error of 23.06\%. With stochastic depth, we obtain an error of 23.38\% at epoch 90, which is slightly higher. We observe from Fig.\ref{figure.imagenet} that the downward trend of the validation error with stochastic depth is still strong, and from our previous experience, could benefit from further training. Due to the 25\% computational saving, we can add 30 epochs (giving 120 in total, after decreasing the learning rate to 1e-4 at epoch 90), and still finish in almost the same total time as 90 epochs of the baseline. This reaches a final error of 21.98\%. We have also kept the baseline running for 30 more epochs. This reaches a final error of 21.78\%. 

Because ImageNet is a very complicated and large dataset, the model complexity required could potentially be much more than that of the 152-layer ResNet~\cite{he2016identity}. In the words of an anonymous reviewer, the current generation of models for ImageNet are still in a different regime from those of CIFAR. Although there seems to be no immediate benefit from applying stochastic depth on this particular architecture, it is possible that stochastic depth will lead to improvements on ImageNet with larger models, which the community might soon be able to train as GPU capacities increase.

\section{Analytic Experiments}

\label{sec_analysis}


In this section, we provide more insights into \name{} by presenting a series of analytical results. We perform experiments to support the hypothesis that \name{} effectively addresses the problem of vanishing gradients in backward propagation. Moreover, we demonstrate the robustness of \name{} with respect to its hyper-parameter.

\begin{figure}[t]
	\vspace{-2ex}
	\begin{center}
		\centerline{\includegraphics[width = 0.9 \columnwidth]{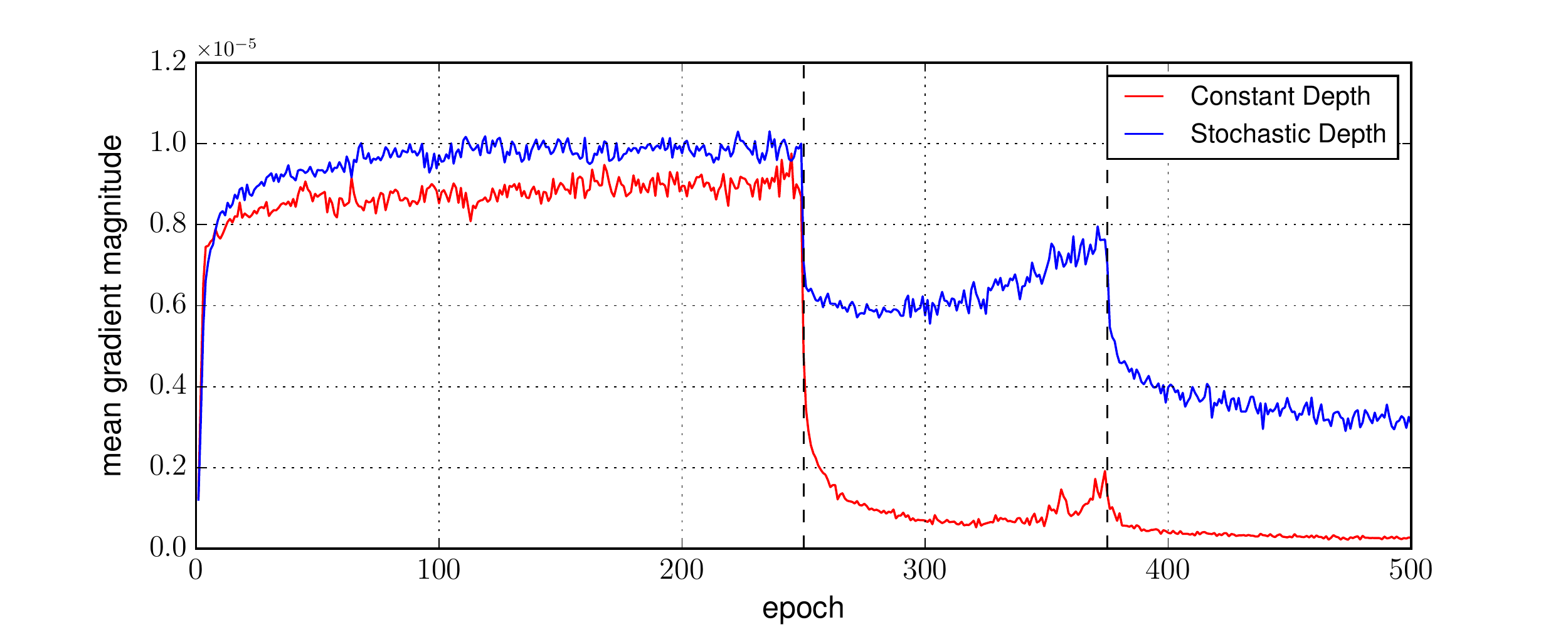}}
		\vspace{-2ex}
		\caption{The first convolutional layer's mean gradient magnitude for each epoch during training. The vertical dotted lines indicate scheduled reductions in learning rate by a factor of 10, which cause gradients to shrink.}
		\label{figure.gradient}
	\end{center}
	\vspace{-8ex}
\end{figure}

\paragraph{\textbf{Improved gradient strength.}}
Stochastically dropping layers during training reduces the effective depth on which gradient back-propagation is performed, while keeping the test-time model depth unmodified. As a result we expect training with \name{} to reduce the vanishing gradient problem in the backward step.
To empirically support this, we compare the magnitude of gradients to the first convolutional layer of the first ResBlock ($\ell\!=\!1$) with and without \name{} on the  CIFAR-10 data set.

Fig. \ref{figure.gradient} shows the mean absolute values of the gradients. The two large drops indicated by vertical dotted lines are due to scheduled learning rate division. It can be observed that the magnitude of gradients in the network trained with \name{} is always larger, especially after the learning rate drops. This seems to support out claim that \name{} indeed significantly reduces the vanishing gradient problem, and enables the network to be trained more effectively.  Another indication of the effect is in the left panel of Fig.~\ref{figure.cifar}, where one can observe that the test error of the ResNets with constant depth approximately plateaus after the first drop of learning rate, while \name{} still improves the performance even after the learning rate drops for the second time. This further supports that \name{} combines the benefits of shortened network during training with those of deep models at test time.

\begin{figure}[t]
	\vspace{-2ex}
	\begin{center}
		\centerline{\includegraphics[width=0.5\columnwidth]{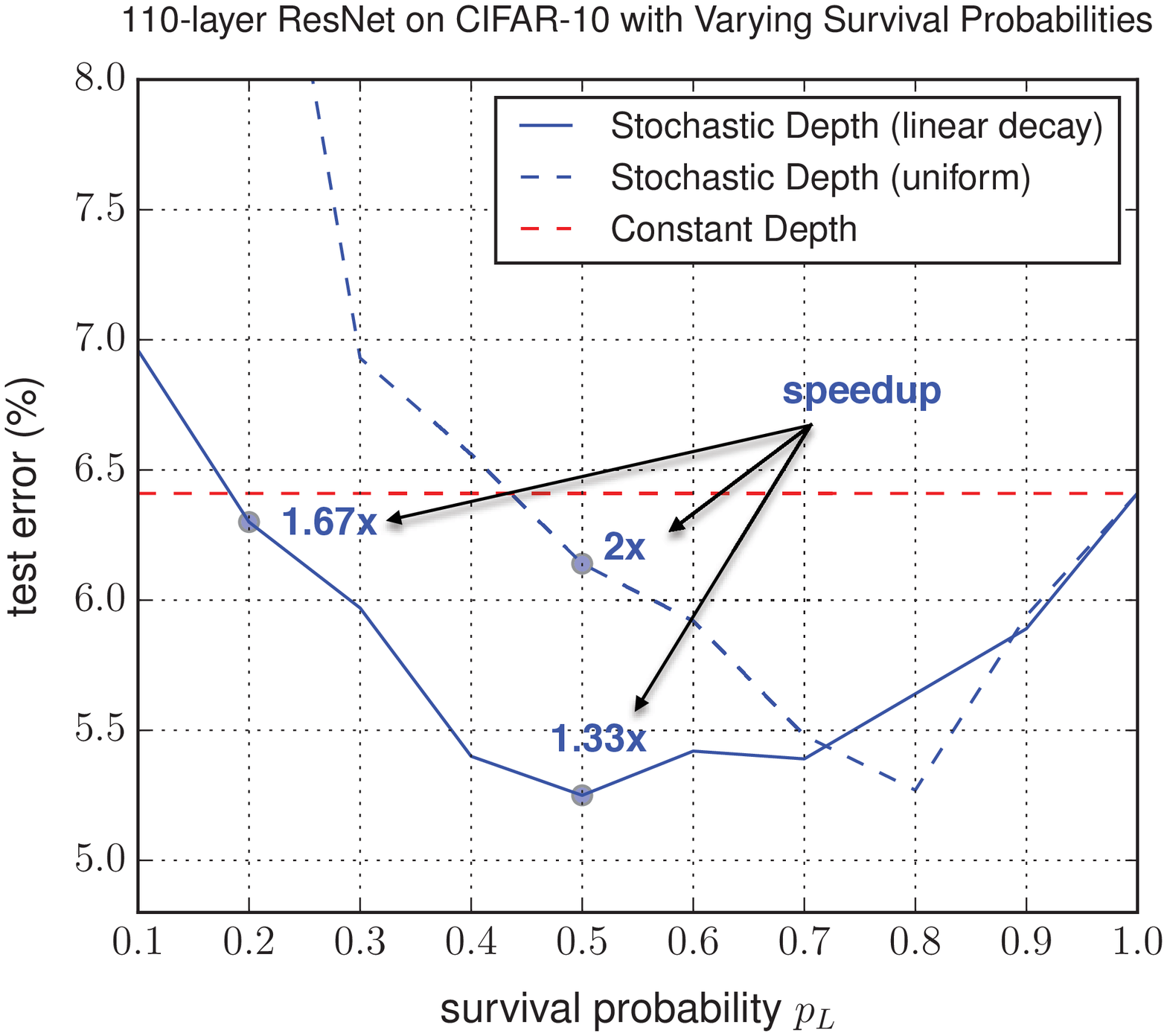}
		\includegraphics[width=0.58\columnwidth]{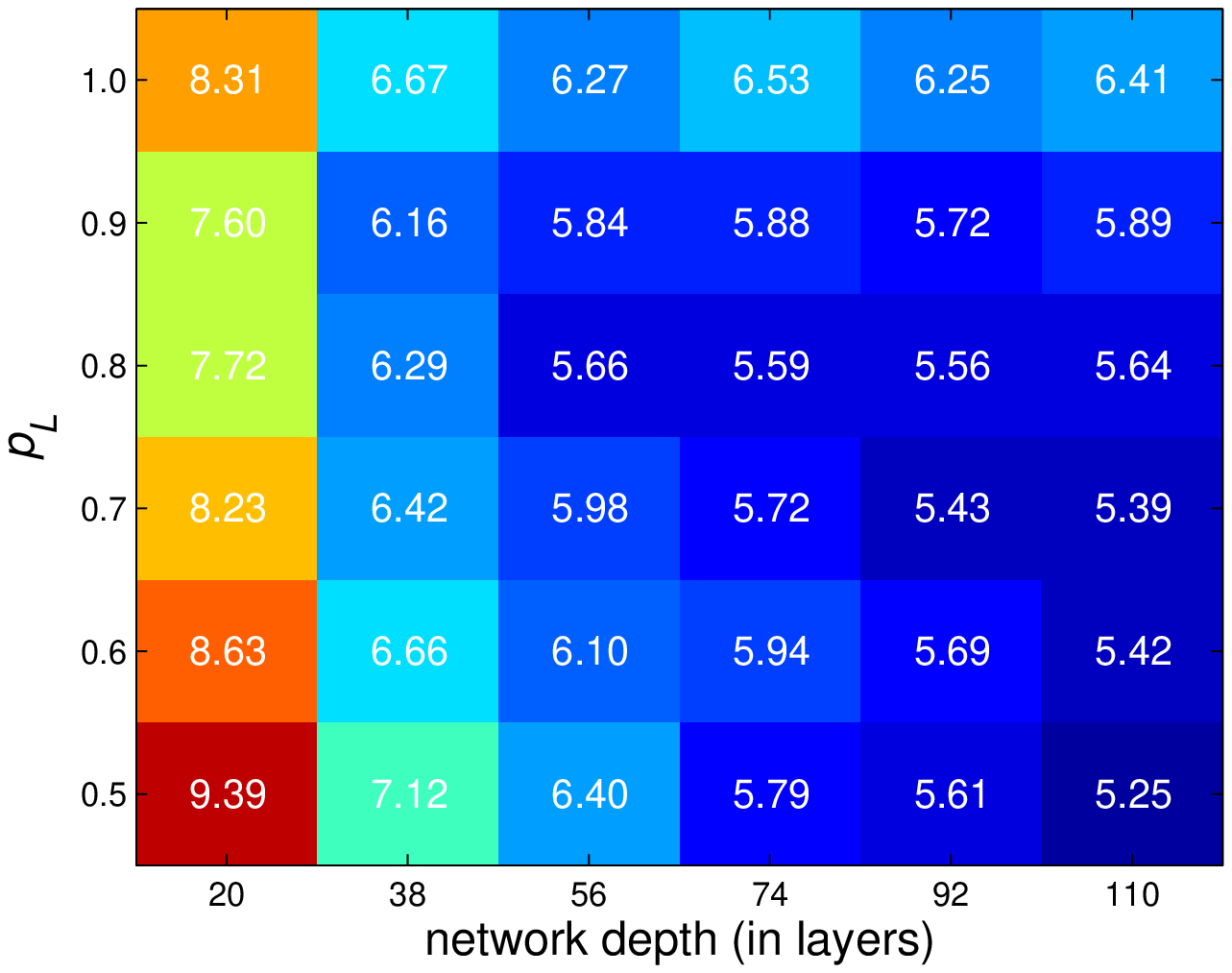}}
	\end{center}
	\vspace{-4ex}
    \label{figure.deathRate}
	\caption{Left: Test error (\%) on CIFAR-10 with respect to the  $p_L$ with uniform and decaying assignments of $p_\ell$. Right: Test error (\%) heatmap on CIFAR-10 varyied over $p_L$ and network depth.}
	\vspace{-4ex}
	\label{figure.deathRate}
\end{figure}

\paragraph{\textbf{Hyper-parameter sensitivity.}}
The survival probability $p_L$ is the only hyper-parameter of our method. Although we used $p_L\!=\!0.5$ throughout all our experiments, it is still worth investigating the sensitivity of \name{} with respect to its hyper-parameter. To this end, we compare the test error of the 110-layer ResNet under varying values of $p_L$ ($L=54$) for both linear decay and uniform assignment rules on the CIFAR-10 data set in Fig.~\ref{figure.deathRate} (\textit{left}). We make the following observations: 1) both assignment rules yield better results than the baseline when $p_L$ is set properly; 2) the linear decay rule outperforms the uniform rule consistently; 3) the linear decay rule is relatively robust to fluctuations in $p_L$ and obtains competitive results when $p_L$ ranges from 0.4 to 0.8; 4) even with a rather small survival probability e.g. $p_L=0.2$, \name{} with linear decay still performs well, while giving a 40\% reduction in training time. This shows that \name{} can save training time substantially without compromising accuracy.

The heatmap on the right shows the test error varied over both $p_L$ and network depth. Not surprisingly, deeper networks (at least in the range of our experiments) do better with a $p_L=0.5$. The "valley" of the heatmap is along the diagonal. A deep enough model is necessary for stochastic depth to significantly outperform the baseline (an observation we also make with the ImageNet data set), although shorter networks can still benefit from less aggressive skipping.

\section{Conclusion}

\label{sec_conclusion}

In this paper we introduced deep networks with \emph{\name{}}, a procedure to train very deep neural networks effectively and efficiently. Stochastic depth reduces the network depth during training \emph{in expectation} while maintaining the full depth at testing time. Training with \name{} allows one to increase the depth of a network well beyond 1000 layers, and still obtain a reduction in test error. Because of its simplicity and practicality we hope that training with \name{} may become a new tool in the deep learning ``toolbox", and will help researchers scale their models to previously unattainable depths and capabilities.

\paragraph{\textbf{Acknowledgements.}} We thank the anonymous reviewers for their kind suggestions. Kilian Weinberger is supported by NFS grants IIS-1550179, IIS-1525919 and EFRI-1137211. Gao Huang is supported by the International Postdoctoral Exchange Fellowship Program of China Postdoctoral Council (No.20150015). Yu Sun is supported by the Cornell University Office of Undergraduate Research. We also thank our lab mates, Matthew Kusner and Shuang Li for useful and interesting discussions.

\bibliographystyle{splncs}

\bibliography{egbib}

\end{document}